\RequirePackage{amsmath}
\documentclass[runningheads]{llncs}
\usepackage{graphicx}
\usepackage{comment}
\usepackage{amsmath,amssymb} 
\usepackage{color}
\usepackage[skip=2pt]{caption}

\usepackage{graphics}
\usepackage{xspace}
\usepackage{color}
\usepackage{bm}

\usepackage{siunitx}
\usepackage{balance}
\usepackage{lmodern}
\usepackage{tabularx}
\usepackage{booktabs}
\usepackage{makecell}
\usepackage[hyphens]{url}
\usepackage{hyperref}

\usepackage{wrapfig}

\newcolumntype{P}[1]{>{\raggedright\arraybackslash}m{#1}}
\newcolumntype{C}[1]{>{\centering\arraybackslash}m{#1}}
\newcolumntype{R}[1]{>{\raggedleft\arraybackslash}m{#1}}

\usepackage{xcolor}

\definecolor{xucongcolor}{rgb}{0.73725, 0.6588, 0.0705}

\usepackage{gensymb}
\usepackage{diagbox}

\newcommand{\datasetname}{ETH-XGaze\xspace}

\newcommand{\numparticipant}{110\xspace}
\newcommand{\numsample}{1,083,492\xspace}

\begin{document}
\pagestyle{headings}
\mainmatter
\def\ECCVSubNumber{3477}  
\title{\datasetname: A Large Scale Dataset \\ for Gaze Estimation under Extreme Head Pose and Gaze Variation}

\titlerunning{ETH-XGaze}
\author{Xucong Zhang\inst{1},
Seonwook Park\inst{1},
Thabo Beeler\inst{2},
Derek Bradley,
Siyu Tang\inst{1} \and
Otmar Hilliges\inst{1}}

\authorrunning{X. Zhang et al.}
\institute{Department of Computer Science, ETH Zurich\\
\email{\{xucong.zhang, spark, siyu.tang, otmar.hilliges\}@inf.ethz.ch}\\
\and
Google Inc.\\
\email{tbeeler@google.com}}

\maketitle

\begin{abstract}
Gaze estimation is a fundamental task in many applications of computer vision, human computer interaction and robotics. Many state-of-the-art methods are trained and tested on custom datasets, making comparison across methods challenging.  Furthermore, existing gaze estimation datasets have limited head pose and gaze variations, and the evaluations are conducted using different protocols and metrics.
In this paper, we propose a new gaze estimation dataset called \datasetname, consisting of over one million high-resolution images of varying gaze under extreme head poses. We collect this dataset from \numparticipant participants with a custom hardware setup including 18 digital SLR cameras and adjustable illumination conditions, and a calibrated system to record ground truth gaze targets.
We show that our dataset can significantly improve the robustness of gaze estimation methods across different head poses and gaze angles.  Additionally, we define a standardized experimental protocol and evaluation metric on~\datasetname, to better unify gaze estimation research going forward. The dataset and benchmark website are available at
\newcommand{\dataseturl}{\url{https://ait.ethz.ch/projects/2020/ETH-XGaze}}
\dataseturl \end{abstract}

\section{Introduction}
Estimating eye-gaze from monocular images alone has recently received significant interest in computer vision~\cite{wang2019generalizing,yu2019improving,fischer2018rt} due to its significance in many application domains ranging from the cognitive sciences and HCI to robotics and semi-autonomous driving~\cite{demiris2007prediction,majaranta2014eye,park2013predicting}. 
Many arising computing paradigms such as smart-home appliances, autonomous cars and robots, as well as body-worn cameras will rely on understanding the attention and intent of humans without directly interacting with the observed person.
We argue that in order  
to be more robust to a larger variety of environmental conditions, future methods should be able to accurately estimate the gaze of humans in a broader range of settings, including variation of viewpoint, extreme gaze angles, lighting variation, input image resolutions, and in the presence of occluders such as glasses. 

Unfortunately, existing gaze datasets do not cater to such use-cases and are mostly limited to the frontal setting, covering a relatively narrow range of head poses and gaze directions. These are typically collected via laptops~\cite{zhang2019mpiigaze}, mobile devices~\cite{krafka2016eye,huang2017tabletgaze} or in stationary settings~\cite{funes2014eyediap}. Recent work has moved towards more unconstrained environmental conditions in particular with respect to lighting but the coverage of head pose and gaze direction ranges remains limited~\cite{kellnhofer2019gaze360,fischer2018rt,yu2020humbi}.

\begin{figure}[t]
    \centering
    \includegraphics[width=0.9\textwidth]{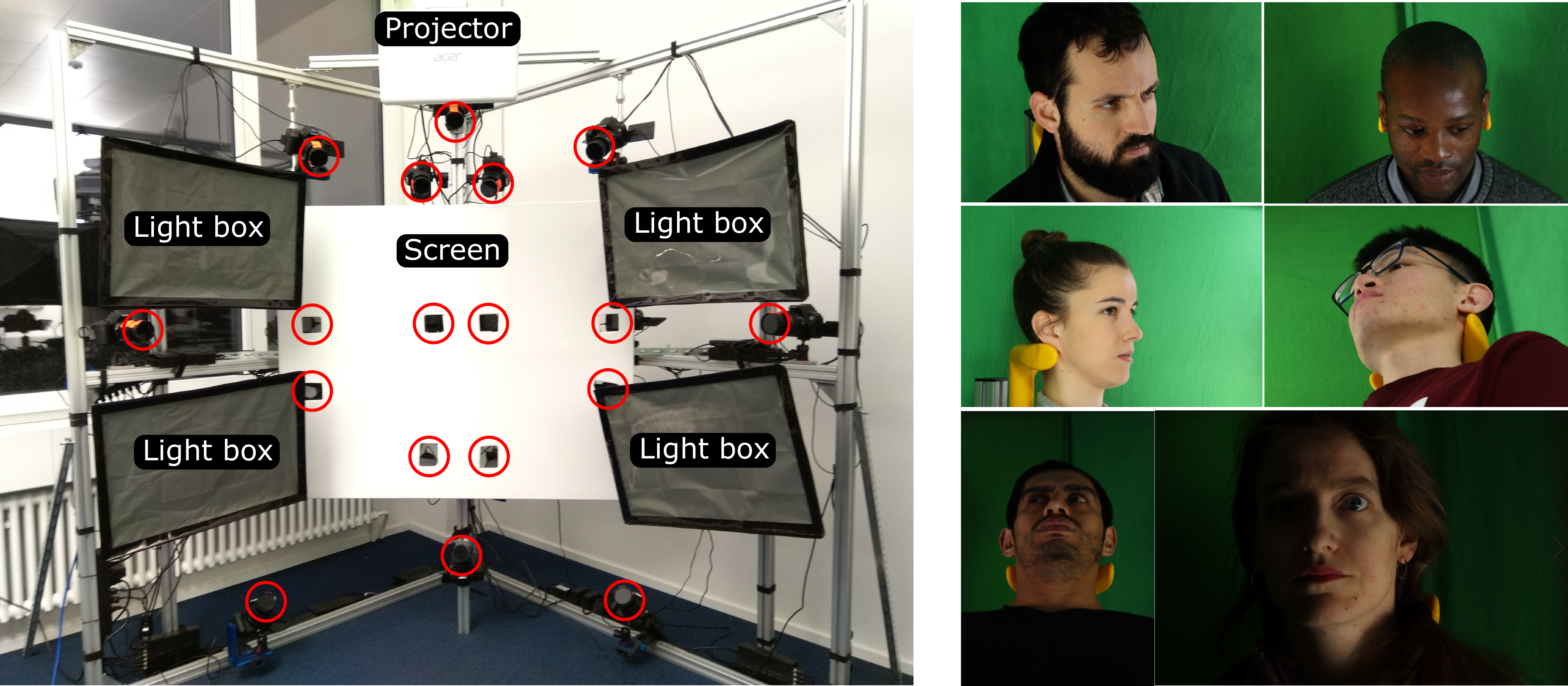}
    \caption{Our data collection device includes 18 high-resolution Canon 250D digital SLR cameras (marked with red circles), a projector to project the stimuli on the screen, and four Walimex Daylight 250 light boxes. A chair with a head rest is positioned approximately one meter away from the screen. Captured samples under different head poses and lighting conditions are shown on the right.}
    \label{fig:device}
\end{figure}

In this paper we detail a new dataset, dubbed~\datasetname, to facilitate research into robust gaze estimation methods. The dataset exhaustively samples large variations in head poses, up to the limit of where both eyes are still visible (maximum $\pm70^{\circ}$ from directly facing the camera) as well as comprehensive gaze directions (maximum $\pm50^{\circ}$ in the head coordinate system)~\cite{ruch1960medical}.
The dataset will allow for the development of new methods that can robustly estimate gaze direction without requiring a quasi-frontal camera placement. We show experimentally that 
i) the data distribution of \datasetname is more comprehensive than other datasets 
(e.g., our dataset broadens the scope for eye-gaze research),
and ii) that training on our dataset significantly improves robustness towards head pose and gaze direction variations. 
Beyond extending the gaze and head-pose ranges, the proposed dataset allocates considerably more pixels to the periocular region compared to existing datasets (e.g. refer to Fig.~\ref{fig:data_sample}). This allows to train gaze estimators that can take advantage of the high-resolution imagery of modern camera hardware to improve gaze prediction. We collect data from~\numparticipant participants with different ethnicity, age, and gender \textendash~ some with glasses and some without \textendash~in order to provide a rich and diverse dataset. For each of the participants we capture over 500 gaze directions with full-on illumination, plus an additional 90 samples under 15 different illumination conditions. This results in a total of over 1 million labeled samples. For all samples, the ground-truth gaze direction is known since the gaze is guided by stimuli displayed on a large screen in front of the participant, ensuring good label quality even under extreme view angles.
The capture setup is depicted in Fig.~\ref{fig:device} (left).

To ensure fair and systematic comparisons between future methods that leverage this new large-scale dataset, we also propose a standardized evaluation protocol. Unlike other fields in computer vision that have benefited from such benchmark frameworks (i.e. image classification~\cite{russakovsky2015imagenet}, face recognition~\cite{nech2017level}, full-body~\cite{h36m}, hand pose estimation~\cite{zimmermann2019freihand} and multiview stereo reconstruction~\cite{seitz2006comparison}), the gaze estimation community has so far relied on a heterogeneous environment where many papers employ custom data pre-processing and evaluation protocols, rendering direct comparisons challenging.
Motivated by the benchmarking approaches in adjacent areas we create a website open to the public to submit, evaluate and compare gaze estimation methods based on \datasetname. 

Finally, in order to provide initial insights into the value of our dataset, we provide results from a simple gaze estimation method that can serve as a baseline. Our estimation approach leverages a standard CNN architecture (i.e., ResNet-50~\cite{he2016deep}), trained with the task of estimating gaze from a monocular face patch. We present the estimation results as well as an ablation study of training on different subsets of our dataset, indicating the importance of all sampled dimensions (e.g. head pose and gaze angles, number of subjects, lighting conditions and input image resolution). We hope this baseline method and evaluations will inspire future research in gaze estimation using our \datasetname dataset.  

\noindent In summary, our contribution is three-fold:
\begin{itemize}
\item A large scale dataset (over 1 Mio samples) for gaze estimation covering a large head pose and gaze range from~\numparticipant participants of different age, gender and ethnicity with consistent label quality and high-resolution images.
\item Standardized experimental protocol and evaluation metrics including a new robustness evaluation.
\item Detailed analysis on different factors for gaze estimation training.
\end{itemize}

\section{Related Work}

\subsection{Gaze Estimation Algorithms}
Initial learning-based gaze estimation methods often assume a static head pose \cite{baluja1994non,lu2011inferring},
with later works allowing for gradually more head pose freedom \cite{lu2014adaptive,sugano2014learning}.
In parallel, gaze estimation errors on public datasets have improved rapidly in recent years, through the use of domain adaptation \cite{shrivastava2017learning}, Bayesian networks \cite{Wang_2018_CVPR}, adversarial approaches \cite{wang2019generalizing}, coarse-to-fine \cite{cheng2020coarse}, and multi-region CNNs \cite{krafka2016eye,fischer2018rt}.
Recent development in the person-specific adaptation of gaze estimators \cite{Park2019ICCV,yu2019improving,liu2018differential} are quickly reducing error metrics on public datasets even further. However, gaze-estimation is studied mostly in the frontal setting which does not apply to many emerging application domains.  
There is hence a need for a systematic method to understanding the robustness of a model with regards to gaze direction and head orientation ranges.
We thus propose our gaze estimation dataset to cover these factors and propose concrete tasks for their evaluation.

\subsection{Gaze Datasets}

Newly introduced datasets in any area of research tend to push the limits of the data distribution represented in existing datasets.
Multi-view cameras have been used to cover head poses in previous works.
However, there are limited range of head poses~\cite{smith2013gaze}, or limited effective resolution on face region using machine vision cameras~\cite{sugano2014learning} or wide-angle cameras~\cite{yu2020humbi}.
The Columbia dataset uses five high-resolution camera while only 5,880 samples with discrete gaze directions are recorded~\cite{smith2013gaze}.
UT Multi-view (UTMV)~\cite{sugano2014learning} is recorded with eight machine version cameras, and HUMBI is recorded with multiple wide-angle cameras, however, their resolution of eye region is small in the captured image.
Capturing different head poses with a single camera can be achieved by asking participants to explicitly move their head during recording as in EYEDIAP~\cite{funes2014eyediap}, moving the camera and gaze target around the participant as in Gaze360~\cite{kellnhofer2019gaze360}, or both as in RT-GENE~\cite{fischer2018rt}.
Some of these approaches do result in lower resolution images, and as such are not informative in the development of generative models \cite{shrivastava2017learning} or gaze redirection methods~\cite{He_2019_ICCV,yu2019improving}. Therefore, these methods had to revert to the synthetic data from UnityEye~\cite{wood2016_etra} or the relatively small Columbia datasets~\cite{smith2013gaze}.
In addition, it is more challenging to aim for the acquisition of a balanced dataset in terms of head pose and gaze estimation ranges when capturing in the wild (cf. \cite{zhang2019mpiigaze,krafka2016eye,kellnhofer2019gaze360}), as is later shown in this paper in parameter range comparisons between our proposed dataset and existing ones.
Our high resolution dataset tackles the mentioned challenge of limited head pose and gaze direction ranges in existing datasets, taking meaningful steps towards constructing a balanced set of training data for learning high performance and robust gaze estimation models. Furthermore, we see potential in leveraging the high quality imagery to enable future work in areas adjacent to gaze-estimation such as generative modeling of the eye-region, Computer Graphics and facial reconstruction.

A comprehensive summary of current gaze estimation datasets in relationship to ours is shown in Tab.~\ref{tab:other_datasets}. 

\subsection{Evaluation Protocols}
Having public benchmark frameworks for evaluation of popular algorithms is common for many computer vision tasks such as image classification~\cite{russakovsky2015imagenet}, face recognition~\cite{kemelmacher2016megaface}, pedestrian detection~\cite{dollar2011pedestrian} and hand pose estimation~\cite{zimmermann2019freihand}.
Unfortunately, there is neither a unified evaluation protocol for gaze estimation nor an existing dataset that can serve as a general evaluation platform.
Despite existing best practices, most previous work relies on their own data pre-processing and sometimes uses different training-test splits for evaluation. 
To provide a platform for gaze estimation evaluation, we share our dataset~\datasetname and define a set of clearly defined evaluation procedures. Furthermore, an online evaluation system and public leader-board are released along with the dataset).  
\section{\datasetname Dataset}

\begin{table}[t]
\centering
\begin{tabularx}{\textwidth}{P{2.1cm} C{1.5cm} C{2.2cm} C{2.2cm} C{1.8cm} C{1.9cm}}
\toprule
& \textbf{\# Peo.} & \textbf{Maximum Head Pose} & \textbf{Maximum Gaze} & \textbf{\# Data} & \textbf{Resolution} \\
\midrule
Columbia~\cite{smith2013gaze} & 56 & $0^{\circ}$, $\pm30^{\circ}$ & $\pm15^{\circ}$, $\pm10^{\circ}$ & 5,880 & $5184\times$3456 \\
UTMV~\cite{sugano2014learning} & 50 & $\pm36^{\circ}$, $\pm36^{\circ}$ & $\pm50^{\circ}$, $\pm36^{\circ}$ & 64,000 & $1280\times1024$\\
EYEDIAP~\cite{funes2014eyediap} & 16 & $\pm15^{\circ}$, $30^{\circ}$ & $\pm25^{\circ}$, $20^{\circ}$ & 237 min & HD \& VGA \\
MPIIGaze~\cite{zhang2019mpiigaze} & 15 & $\pm15^{\circ}$, $30^{\circ}$ & $\pm20^{\circ}$, $\pm20^{\circ}$ & 213,659 & $1280\times720$ \\
GazeCapture~\cite{krafka2016eye} & 1,474 & $\pm30^{\circ}$, $40^{\circ}$ & $\pm20^{\circ}$, $\pm20^{\circ}$ & 2,445,504 & $640\times480$\\
RT-GENE~\cite{fischer2018rt} & 15 & $\pm40^{\circ}$, $\pm40^{\circ}$ & $\pm40^{\circ}$, $-40^{\circ}$ & 122,531 & $1920\times1080$\\
Gaze360~\cite{kellnhofer2019gaze360} & 238 & $\pm90^{\circ}$, unknown & $\pm140^{\circ}$, $-50^{\circ}$ & 172,000 & $4096\times3382$ \\
\midrule
\textbf{\datasetname} & \numparticipant & $\bm{\pm80^{\circ}}$, $\bm{\pm80^{\circ}}$ & $\bm{\pm120^{\circ}}$, $\bm{\pm70^{\circ}}$ & \numsample & $\bm{6000\times4000}$ \\
\bottomrule
\end{tabularx}
\caption{Overview of popular gaze estimation datasets showing the number of participants, the maximum head poses and gaze in horizontal (around yaw axis) and vertical (around pitch axis) directions in the camera coordinate system, amount of data (number of images or duration of video), and image resolution.}
\label{tab:other_datasets}
\end{table}

\noindent There are several parameters that define a comprehensive gaze estimation dataset, including: head pose, gaze direction, subject appearance, illumination condition, and image resolution.
We design the~\datasetname data collection procedure with the main objective to maximize the parameter range along each of those dimensions as much as possible.

\subsection{Acquisition Setup}
The setup used for data collection is shown in the left of Fig.~\ref{fig:device}.
We capture the subject with 18 Canon 250D digital SLR cameras from different viewpoints to cover a large range of head poses.
There are five paired cameras for geometry capture and eight cameras for texture acquisition, such as to enable 3D face reconstruction in the future. The resolution of the captured images is very high ($6000\times4000$ pixels).
All cameras are connected via ESPER trigger boxes\footnote{https://www.esperhq.com} to a Raspberry Pi, and a wireless mouse is used to send the triggering signal to the Raspberry Pi.
The delay between mouse click and triggering the camera is below 0.05 seconds.
A large screen ($120\times100$ cm) is placed in the center of the cameras to show the stimuli controlled by the Raspberry Pi and projected by a projector. Since some cameras are placed behind the screen, we create cutout holes for their lenses. 
There are four light boxes (Walimex Daylight 250) surrounding the screen and each of them is equipped with a light bulb that emits $\sim$4500lm.
The Raspberry Pi can turn each of the light boxes on or off to simulate different illumination conditions.
We mount polarization filters in front of both the light box and camera and carefully adjust the filter angle to attenuate specular reflection off the face of the participants.
During recording, the participants are sitting at approximately one meter distance in front of the screen, with the head placed in a head rest to reduce unintentional head motion.

\begin{figure}[t]
    \centering
    \includegraphics[width=\textwidth]{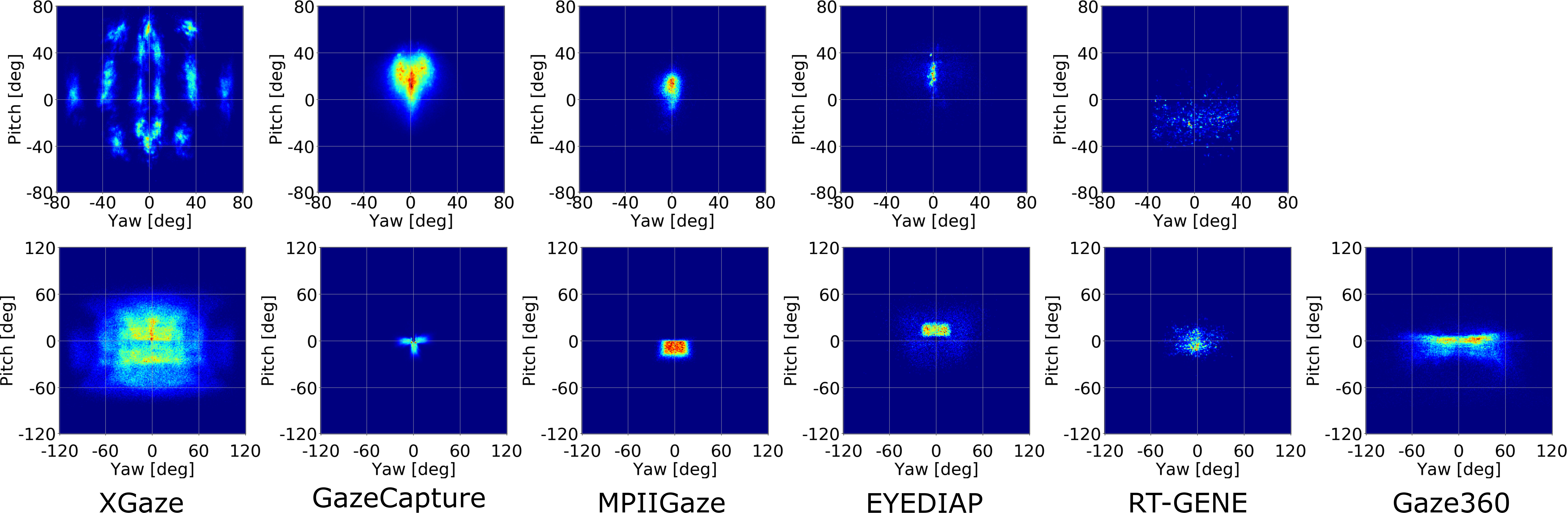}
    \caption{Head pose (top row) and gaze direction (bottom row) distributions of different datasets. The head pose of Gaze360 is not shown here since it is not provided by the dataset.}
    \label{fig:head_gaze_dis}
\end{figure}
\subsection{Collection Procedure}
During data collection, the participant focuses on a shrinking circle and clicks the mouse when the circle becomes a dot, providing the gaze point.
The position of gaze points are randomly distributed on the screen.
We have three methods to ensure the participant is looking at the dot when clicking the mouse.
First, the participant has a short time window of 0.5 second to click the mouse to successfully collect one sample.
Second, the shrinking time of the circle is random such that the participant has to focus on the shrinking circle to avoid missing the triggering time window.
Third, the participant is told to collect a fixed amount of samples and any missing mouse click will increase the collection time.
For most of the data collection, all four light boxes are fully on, in order to provide the maximum brightness, but we additionally simulate 15 illumination conditions by switching on and off the four light boxes.

\subsection{Data Characteristics}

\begin{figure}[t]
    \centering
    \includegraphics[width=0.95\textwidth]{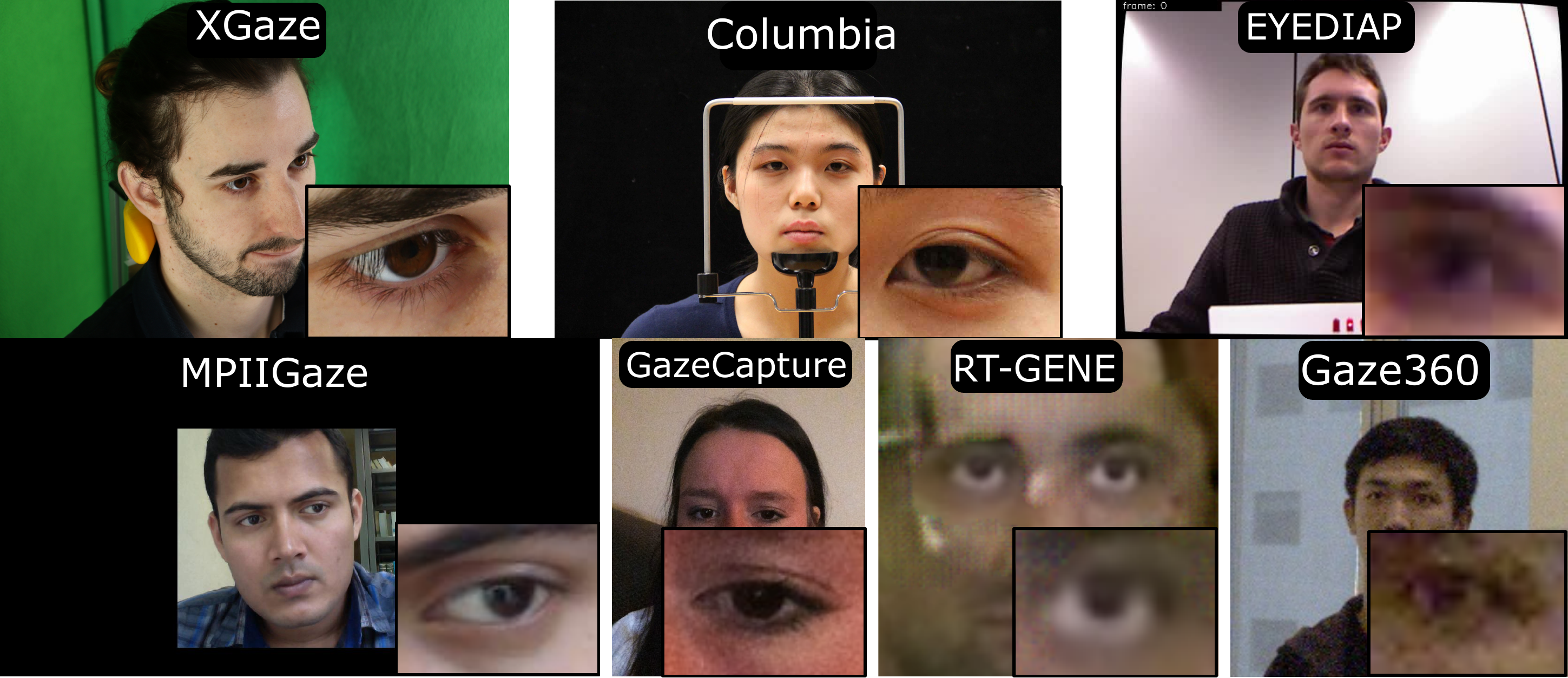}
    \caption{Data examples and corresponding cropped eye images from different gaze estimation datasets. \datasetname has the highest image resolution and quality.}
    \label{fig:data_sample}
\end{figure}

In total, we collect data from 110 participants (47 female and 63 male), aged between 19 and 41 years.
17 of them wore contact lenses and 17 of them wore eye glasses during recording.
The ethnicities of the participants includes Caucasian, Middle Eastern, East Asian, South Asian and African.
Each participant collected 525 gaze points under the full-lighting condition, and 90 gaze points under the varying lighting conditions - six gaze points for each of the 15 lighting conditions.
For each gaze point, a total of 18 images was collected by the 18 different cameras.
We manually removed samples for which the participant was not looking at the ground-truth point-of-regard due to  blinking, motion blur etc.
This results in total~\numsample images samples for whole~\datasetname dataset.

A comparison between the proposed and existing datasets can be found in Tab.~\ref{tab:other_datasets}.
Our dataset surpasses existing datasets regarding the following aspects.

\textbf{Head pose.} \datasetname has the largest range of head poses compared to existing datasets, as shown in the first row of Fig.~\ref{fig:head_gaze_dis}. Examples from~\datasetname with different head poses are shown in Fig.~\ref{fig:head_pose_sample}.
In \cite{kellnhofer2019gaze360}, it is stated that the effective head pose range of Gaze360 is $\pm90^{\circ}$ in horizontal direction and limited head poses in vertical direction. However, head pose annotations are not provided in their dataset and hence we cannot visualize it here.

\textbf{Gaze direction.} \datasetname has the largest range of gaze directions compared to existing datasets.
The second row of Fig.~\ref{fig:head_gaze_dis} compares the gaze direction distributions.
Although Gaze360 reports $\pm140^{\circ}$ coverage on the horizontal gaze direction, the dataset contains only very few samples beyond $\pm70^{\circ}$.
\datasetname is evenly sampled across a large range of horizontal and vertical gaze directions.

\textbf{Image resolution.} \datasetname has the highest image resolution compared to existing datasets, especially the effective resolution on the face region.
We show some examples and corresponding cropped eye images from different datasets in Fig~\ref{fig:data_sample}.
The Columbia dataset also has high image resolution, however, the dataset is comprised of only 5,880 samples.
While EYEDIAP, MPIIGaze, RT-GENE and Gaze360 have fairly high resolution imagery as well, the participant is far away from the camera which results in low effective eye region resolution.

\begin{figure}[t]
    \centering
    \includegraphics[width=1.0\textwidth]{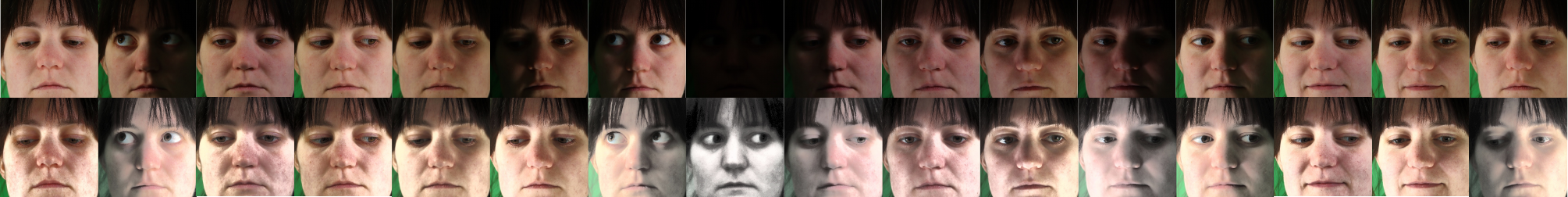}
    \caption{Samples of the 16 illumination conditions created by switching on and off the four light boxes. The first row are the original samples, and the second row employs histogram equalization. The first column is the full-lighting setting.}
    \label{fig:lighting_sample}
\end{figure}

\textbf{Controlled illumination conditions.} \datasetname provides a set of controlled illumination conditions.
Although uncontrolled in-the-wild illumination conditions are important for gaze estimation~\cite{zhang2019mpiigaze,krafka2016eye}, controlled illumination conditions provide complementary information to better understand illumination impact and enable light synthesis.
As shown in Fig.~\ref{fig:lighting_sample}, we record 16 different illumination conditions.

\subsection{\datasetname Utility}
\datasetname makes it possible to \emph{train} gaze estimators that cover large ranges of head poses and gaze directions. This allows to better estimate gaze from oblique viewpoints, such as overhead cameras.
\datasetname can also be used to \emph{evaluate} the robustness of a gaze estimation method with respect to these factors.
In our dataset the head pose remains fixed and thus does not follow the traditional \emph{head-pose-following-gaze} pattern. However, by imaging from 18 viewpoints we densely sample all natural pose-gaze combinations with respect to the camera, suitable for varied applications like gaze estimation from a personal laptop or attention measurement inside a smart home.

Our dataset allows future gaze prediction methods to train on high-resolution imagery, which is critical for generative methods~\cite{He_2019_ICCV,Park2019ICCV,yu2019improving,Wang_2018_CVPR}.
Since the generated image quality highly depends on the training image quality, Columbia and the synthetic UnityEYE dataset have been used during training in the past. 
Our~\datasetname provides high-resolution images ($6000\times4000$ pixels), and more importantly the face region occupies a large portion of the image.

Since the data in \datasetname has been captured to allow for 3D geometry reconstruction using multi-view photogrammetry methods (i.e. \cite{beeler2010high}), it provides the potential of synthesizing high-quality gaze estimation data in the future. Parametric eye models \cite{wood20163d,berard2016lightweight} can be fit to the data to build a controllable rig of the eye \cite{berard2019practical}.
Such a rig can then be used to re-render novel images of different lighting conditions, gaze directions, and head poses with state-of-the-art rendering techniques, providing additional training data for gaze estimation task.

\subsection{Data Pre-processing}
\label{sec:data_preprocess}

\begin{figure}[t]
    \centering
    \includegraphics[width=0.9\textwidth]{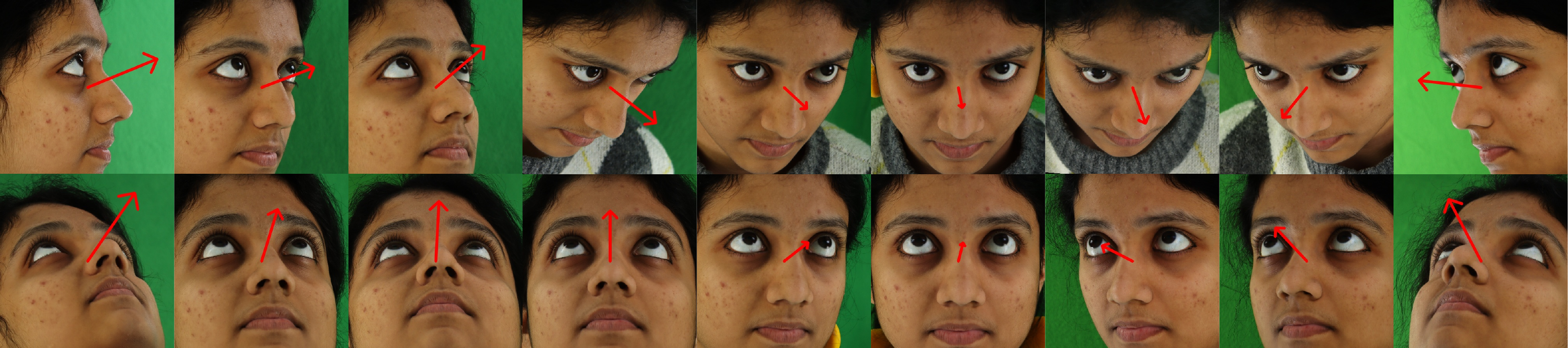}
    \caption{Data examples captured by 18 different camera views. The red arrow is the gaze direction. The face patch images shown are after data normalization.}
    \label{fig:head_pose_sample}
\end{figure}

We crop the face patch out of the original image as input for gaze estimation model training.
For each input image sample, we first perform face and facial landmark detection using a state-of-the-art method~\cite{bulat2017far}.
We then fit a 3D morphable model of the face to the detected landmarks to estimate the 3D head pose~\cite{huber2016multiresolution}.
The 3D head pose along with camera calibration information is used to perform data normalization~\cite{zhang2018revisiting}.
In a nutshell, the data normalization method maps the input image to a normalized space where a virtual camera is used to warp the face patch out of the original input image according to 3D head pose.
It rotates a virtual camera to cancel the head rotation around the row axis, and moves the virtual camera to a fixed distance from the face center to warp the face patch of fixed size.
More details can be found in the original paper~\cite{zhang2018revisiting}.
During data normalization, we define the face center as the center of the four eye corners and two nose corners, we set the focal length of the virtual camera to be 960 mm, the normalized distance to be 300 mm, and the cropped face image is $448\times448$ pixels.
Examples of face patches after data normalization are shown in Fig.~\ref{fig:head_pose_sample}.
The processed data along with original imagery are released to public.

 \section{Evaluation Protocol}
One goal of this paper is to establish a benchmark to evaluate gaze estimation algorithms.
For this purpose, we define four evaluations on~\datasetname.
The first three evaluations - cross-dataset, within-dataset, and person-specific evaluations - are popular evaluations found in the current gaze estimation literature. In addition, we propose to also assess robustness over head poses and gaze directions as a fourth evaluation criteria, which is made possible by~\datasetname.

\subsection{Baseline Method}
We provide a baseline gaze estimation method using an off-the-shelf ResNet-50 network~\cite{he2016deep}.
This baseline takes the full-face patch covering $224\times224$ pixels as input and outputs the horizontal and vertical gaze angles.
We used the ADAM~\cite{kingma2014adam} optimizer with an initial learning rate of 0.0001, and the batch size is set to be 50.
We trained the baseline model for 25 epochs and decay the learning rate by a factor of 0.1 every 10 epochs.

\subsection{Dataset Preparation}
We split~\datasetname into three parts: a training set $\mathbb{TR}$ comprised of 80 participants, a test set for within-dataset evaluation $\mathbb{TE}$ containing 15 participants, and a test set for person-specific evaluation $\mathbb{TES}$ consisting of another 15 participants. Splitting the test data into two disjoint sets allows us to release ground truth gaze required for the person-specific evaluation (Sec.~\ref{sec:eval_person_specific}).
We ensured that the subjects in both training and test sets exhibit diverse gender, age, and ethnicity, some with and some without glasses. While we release both ground-truth gaze and imagery for the training set, we withhold the ground-truth gaze for the test sets. Authors are encouraged to submit gaze predictions on test samples to the benchmark website, and the performance will be evaluated and reported. This enables future research to compare to existing methods on neutral grounds.

Aside from the proposed~\datasetname dataset, we also evaluated other existing datasets with our baseline method.
These datasets were pre-processed as we described in Sec.~\ref{sec:data_preprocess}. For the \textit{EYEDIAP} dataset, we used both screen sequence and floating target sequences and sampled the video sequences every 15 frames. For the \textit{GazeCapture} dataset, we used the pre-processing pipeline from~\cite{Park2019ICCV} to obtain 3D head poses since the dataset does not provide camera parameters.
For the \textit{Gaze360} dataset, we used the face bounding box provided by the dataset to crop the face patch, alongside the 3D gaze ground-truth.
We will ask authors of these datasets for permission to release the processed data such that the community can use it for evaluations on \datasetname.

\subsection{Cross-dataset Evaluation}
\begin{table}[t]
\begin{center}
\begin{tabular}{| c | c | c | c | c | c || c || c |}
\hline
 \diagbox{Train}{Test} & MPIIGaze & EYEDIAP & \makecell{Gaze\\Capture} & \makecell{RT\\-GENE} & Gaze360 & \datasetname & \makecell{Ave.\\ Rank}\\
 \hline
 MPIIGaze & - & 17.9 & \textbf{6.3} & 14.9 & 31.7 & 34.9 & 2.6\\
 \hline
 EYEDIAP & 16.9 & - & 14.2 & 15.6 & 33.7 & 41.7 & 4.2 \\ 
 \hline
 GazeCapture & \textbf{4.5} & 13.7 & - & \textbf{14.7} & 30.2 & 29.4 & \textbf{1.8} \\
 \hline
 RT-GENE & 12.0 & 21.2 & 13.2 & - & 34.7 & 42.6 & 4.6 \\ 
 \hline
 Gaze360 & 10.3 & 11.3 & 12.9 & 26.6 & - & \textbf{17.0} & 2.8 \\ 
 \hline
 \hline
 \datasetname& 7.5 & \textbf{11.0} & 10.5 & 31.2 & \textbf{27.3} & - & 2.0\\
 \hline
\end{tabular}
\end{center}
\caption{Gaze estimation errors in degrees on cross-dataset evaluations. The last column shows the average ranking on each test sets, and all other numbers are gaze estimation error in degrees.
}
\label{tab:cross_dataset}
\end{table}

Cross-dataset evaluation has gained popularity since it indicates the generalization capabilities of a gaze estimation method.
We define the cross-dataset evaluation as training the model on~\datasetname and testing on other datasets, as well as training on other datasets and testing on~\datasetname.

We conducted the pair-wise cross-dataset evaluations on different datasets and show results achieved by the baseline in Tab.~\ref{tab:cross_dataset}.
The results exhibit rather large gaze estimation errors when testing on our~\datasetname, indicating that there is a big domain gap between~\datasetname and previous datasets.
This stems from the fact that~\datasetname exhibits much larger variation in head pose and gaze direction compared to other datasets.
Therefore, the gaze estimator has to extrapolate to those unseen head poses and gaze directions which is known to be a difficult machine learning task.

Training on GazeCapture achieves the best overall ranking since it contains similar head pose and gaze ranges compared to MPIIGaze, RT-GENE and EYEDIAP.
However, it performs poorly on test datasets that exhibit large variation in head pose and gaze direction such as Gaze360 and our \datasetname.
In contrast, \datasetname enables thorough benchmarking of generalization capabilities of future gaze estimation approaches.

\begin{figure}[t]
    \centering
    \includegraphics[width=0.95\textwidth]{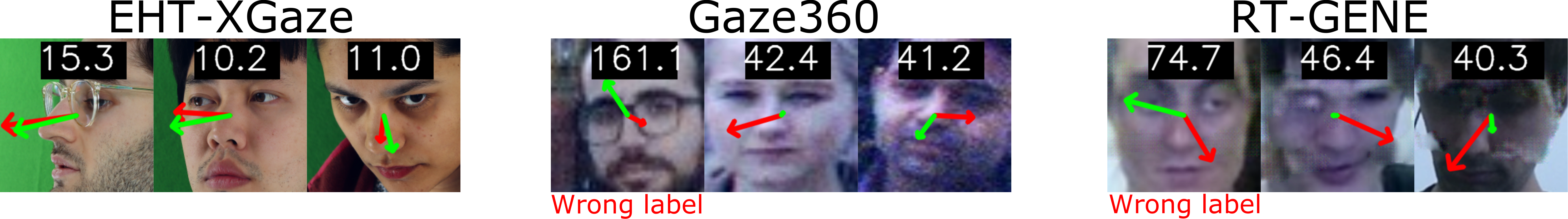}
    \caption{Test samples from different datasets. We show results from training on Gaze360 and testing on~\datasetname (left), and training on~\datasetname and testing on Gaze360 (middle) and RT-GENE (right). The green arrow denotes ground truth and the red arrow is the prediction. The numbers give the respective gaze estimation errors in degrees.}
    \label{fig:test_sample}
\end{figure}

The model trained on Gaze360 achieves the best cross-dataset performance on~\datasetname since they contain similar head pose and gaze direction ranges.
However, Gaze360 has been collected ``in the wild'' setting and can suffer from low-quality images and gaze labels (see Fig.~\ref{fig:test_sample}). 
Our dataset, despite the lab setting, still allows for good performance (the best on EYEDIAP and Gaze360) without any data augmentation.

\subsection{Within-dataset Evaluation}
\label{sec:eval_within_dataset}

\begin{table}[t]
\begin{center}
\begin{tabular}{| c | c | c | c | c | c |}
\hline
  & \datasetname & MPIIGaze & EYEDIAP & GazeCapture & RT-GENE\\
 \hline
 ~\cite{Park2019ICCV} & - & 5.2 & - & 3.5 & -\\
 \hline
 ~\cite{fischer2018rt} & - & 4.8 & - & - & \textbf{8.7} \\
 \hline
 ~\cite{Park2018ECCV} & - & \textbf{4.5} & 10.3 & - & - \\
 \hline
 ~\cite{yu2020unsupervised} & - & - & 6.8 & - & -\\
 \hline
 Baseline & 4.5 & 4.8 & \textbf{6.5} & \textbf{3.3} & 12.0 \\
 \hline
\end{tabular}
\end{center}
\caption{Comparison of the baseline with current state-of-the-art on within dataset evaluations. Numbers are gaze estimation errors in degrees.}
\label{tab:within_dataset}
\end{table}

Within-dataset evaluation is another popular means of evaluating gaze estimation methods.
Here the method is trained on $\mathbb{TR}$ and evaluated on $\mathbb{TE}$.
Tab.~\ref{tab:within_dataset} shows performances of the baseline alongside comparisons to recent state-of-the-art methods. The baseline achieves an error of 4.7 degrees on average on~\datasetname, which is reasonably low given the large ranges of head poses and gaze directions. On other datasets, the baseline exhibits an accuracy comparable to current state-of-the-art methods, indicating that it is a strong baseline. The results of the other methods are taken from the respective publications.

\subsection{Person-specific Evaluation}
\label{sec:eval_person_specific}

\begin{figure}[t]
    \centering
    \includegraphics[width=0.9\textwidth]{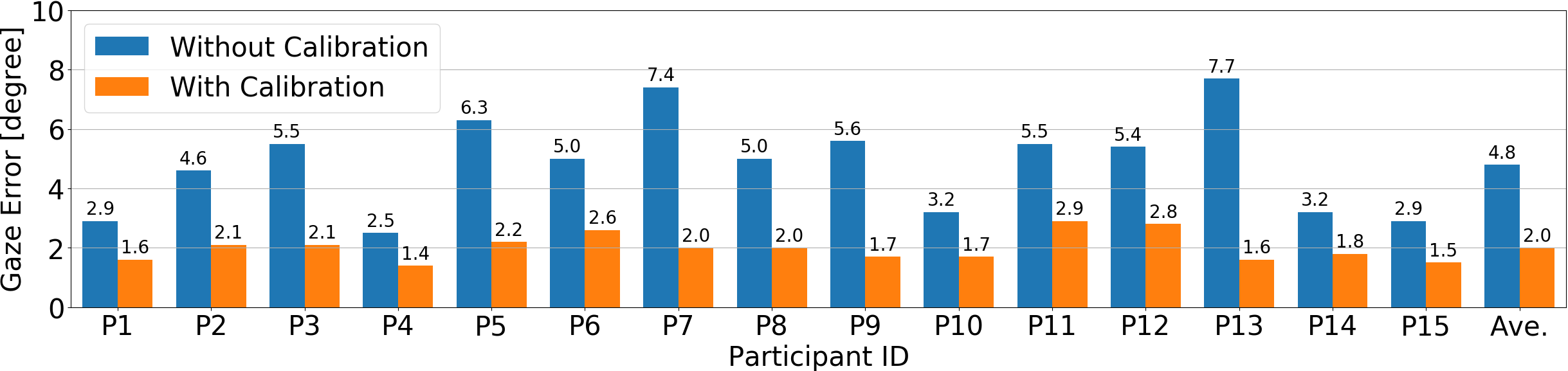}
    \caption{Gaze estimation errors for person-specific evaluation of our baseline. We show the gaze estimation errors with and without training with 200 calibration samples. The number above each bar is the gaze estimation error in degrees.}
    \label{fig:person_specific}
\end{figure}

Person-specific gaze estimation has gained a lot of attention in recent years~\cite{Park2019ICCV,yu2019improving,liu2018differential} due to the huge improvements that can be achieved from even just a few personal calibration samples.
We randomly selected 200 samples from each participant in $\mathbb{TES}$ as the personal calibration samples.
The protocol is to train the model with $\mathbb{TR}$ and up to 200 personal calibration samples, and to test on the remaining samples of $\mathbb{TES}$ -- for each of the 15 test subjects. We pre-trained the model on $\mathbb{TR}$ and then fine-tune it using the 200 samples with 25 epochs.

Results from the baseline in Fig.~\ref{fig:person_specific} show that personal calibration improves the gaze estimates by a large margin.
The goal of this evaluation is not only to achieve good results but also to rely on as few calibration samples as possible.

\subsection{Robustness Evaluation}
\begin{figure}[t]
    \centering
    \includegraphics[width=0.9\textwidth]{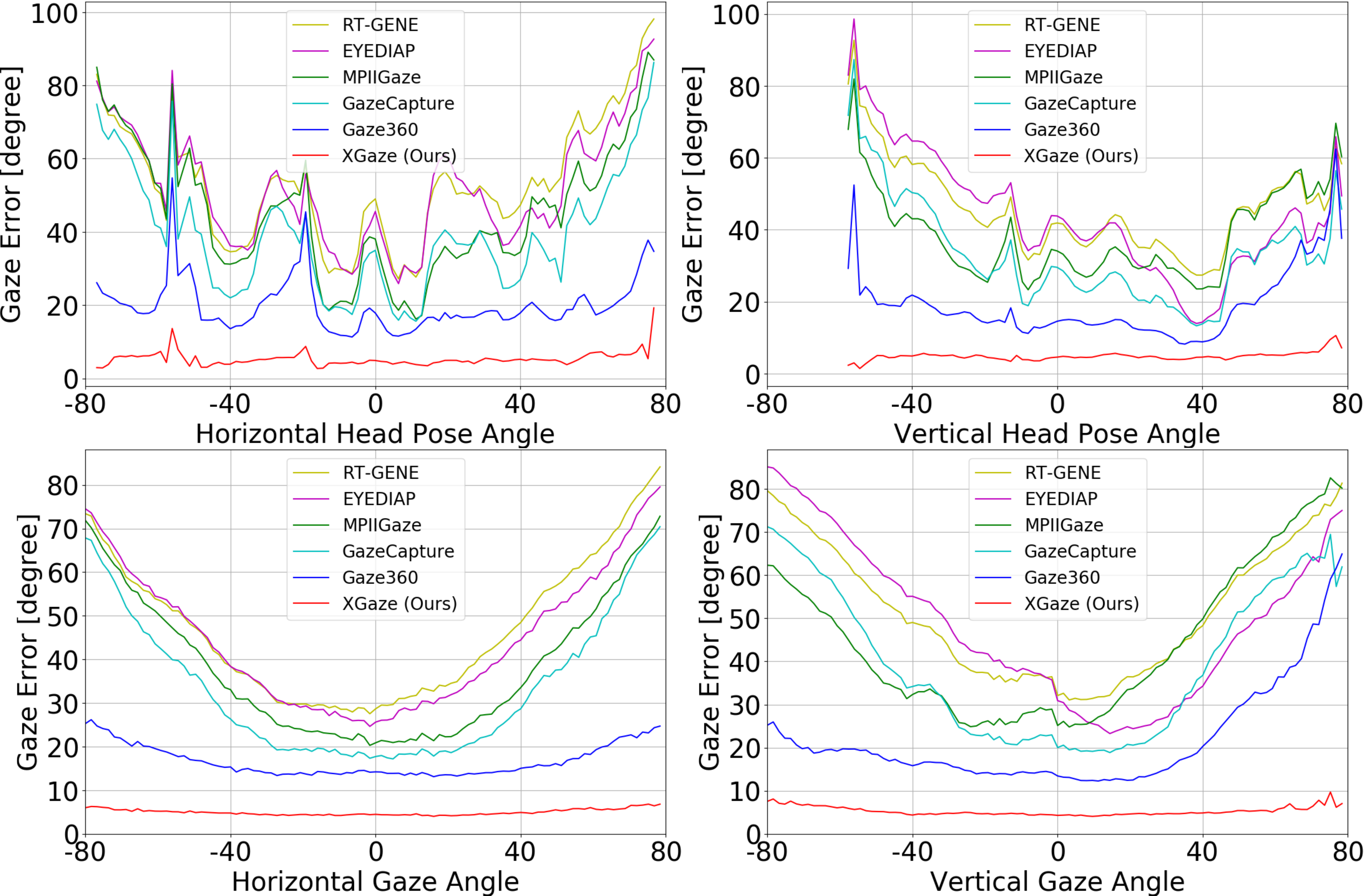}
    \caption{Gaze estimation error distribution across head poses (first row) and gaze directions (second row) in horizontal and vertical directions respectively. The colored curves represent results with different training sets tested on $\mathbb{TE}$.}
    \label{fig:error_dis_eva}
\end{figure}

Previous gaze estimation works usually only report the mean gaze estimation errors without detailed analysis across head poses and gaze directions. This is partly due to the lack of sufficient data samples to cover a wide range. Knowing the performance of an algorithm with respect to these factors is important, since a method with a higher overall error might have lower error within a specific range of interest. Hence we introduce a detailed evaluation to show the robustness across head poses and gaze directions. Fig.~\ref{fig:error_dis_eva} shows the performance of the baseline on $\mathbb{TE}$ over horizontal and vertical axes of the head pose and gaze direction. The different colors represent the different training sets. While these plots evaluate the performance of the different training sets, the benchmark will compare different algorithms instead.
A flat curve across the entire graph, as in the case of training on~\datasetname, indicates robustness to head pose and gaze direction variation.

\section{Demonstration of~\datasetname}

In this section, we evaluate the importance of different factors during training.
Previous gaze estimation datasets cannot serve as the evaluation set for an ablation study of different factors such as head poses, gaze directions and illumination conditions due to the limited coverage.
In contrast, the proposed~\datasetname is an ideal dataset for these evaluations.

\begin{figure}[t]
    \centering
    \includegraphics[width=0.95\textwidth]{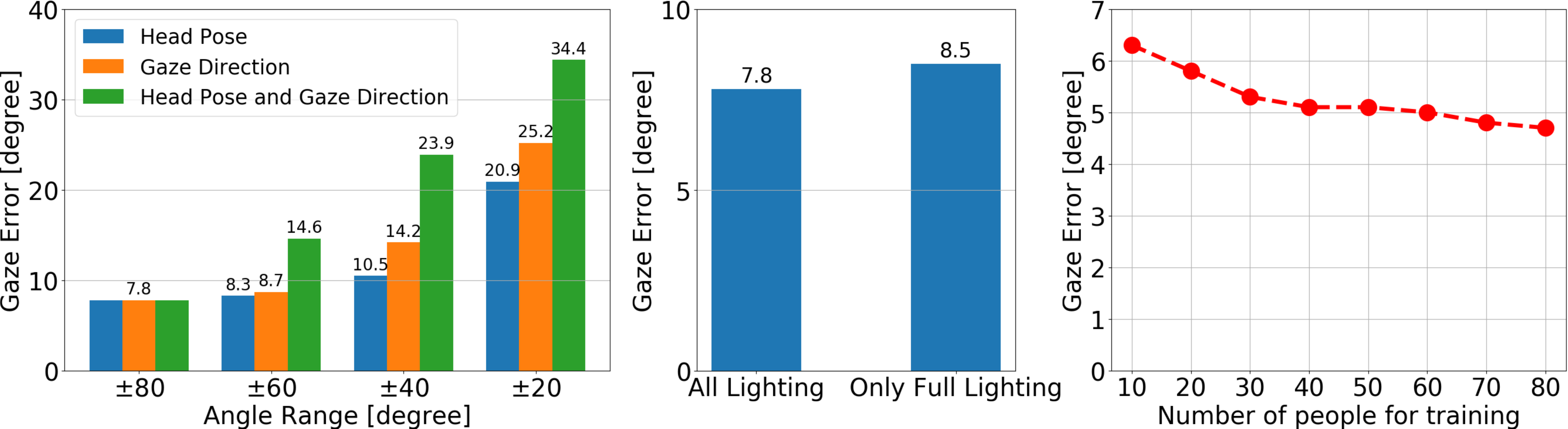}
    \caption{Gaze estimation error distribution by constraining head poses and gaze directions (left), lighting conditions (middle), and number of people (right) during training. The number above each bar is the gaze estimation error in degrees.}
    \label{fig:diff_factors}
\end{figure}

\textbf{Head Pose and Gaze Direction.} We created several training subsets from $\mathbb{TR}$ by constraining the head poses and/or gaze directions angle ranges to be $\pm$80, $\pm$60, $\pm$40, and $\pm$20 in both horizontal and vertical directions.
To keep the same amount of training samples for each subsets, we randomly re-sampled each training subset to have the same amount of samples as the minimal training set, i.e. the training set of $\pm$20 in both head poses and gaze directions.
The results of testing on $\mathbb{TE}$ are shown in the left of Fig.~\ref{fig:diff_factors}.
As we can see from the figure, constraining the head pose and gaze direction results in worse performance in general, especially when we constrain both head pose and gaze direction ranges.
Constraining the gaze directions achieves worse results than constraining head poses, which indicates gaze directions have more impact than the head poses.
Specifically, when we constrained the angle range to be $\pm$40 degrees, the performance decrease caused by constraining head poses is 34.6\%, constraining gaze directions is 82.1\%, and constraining both head poses and gaze directions is 206.4\%.

\textbf{Illumination condition.} In the center of Fig.~\ref{fig:diff_factors}, we show results by training the baseline with all lighting conditions or only with the full-lighting condition.
The performance drop ($9\%$ from 7.8 degrees to 8.5 degrees) indicates the impact of lighting conditions on gaze estimation performance.

\textbf{Personal appearance.} In~\cite{krafka2016eye}, the authors show gaze estimation performance with different numbers of participants.
Our repeated experiment with our baseline on~\datasetname shows the same trend as increasing number of participants improves the performance (see Fig.~\ref{fig:diff_factors}, right).

\textbf{Input resolution.} 
The image resolution analysis in~\cite{zhang2019mpiigaze} was only for eye images and the highest resolution was $60\times36$.
The default input face patch image size to ResNet is $224\times224$ which we used in our baseline.
We resized the input image to be $112\times112$ and $448\times448$ and then fed them into the baseline.
Since there is an average pooling layer at the end of the ResNet convolutional layers, we do not need to modify the architecture with respect to different resolutions.

\begin{table}[t]
\begin{center}
\begin{tabular}{| c | c | c | c |}
\hline
  \diagbox{Train}{Test} & $112\times112$ & $224\times224$ & $448\times448$ \\
 \hline
  $112\times112$ &\textbf{5.4} & 25.3 & 37.2 \\
 \hline
 $224\times224$ & 20.2 & \textbf{4.5} & 42.1 \\
 \hline
 $448\times448$ & 65.1 & 54.7 & \textbf{4.2} \\
 \hline
\end{tabular}
\end{center}
\caption{Gaze estimation errors in degrees generated by models trained with different input image sizes in pixels.}
\label{tab:sizes}
\end{table}

The results of resolution variation are shown in Tab.~\ref{tab:sizes}.
The performance is improved when training and testing on higher resolutions, which indicates the potential of high-resolution gaze estimation.
However, different with results in~\cite{zhang2019mpiigaze}, the model trained on one size achieves much worse results on other sizes.
This can be caused by the much higher image resolution in~\datasetname with large appearance differences compared to the MPIIGaze in~\cite{zhang2019mpiigaze}.
We did not specifically develop the method to handle cross-resolution input images and expect future works can properly deal with cross-resolution training.

 \section{Conclusion}
We present a new large-scale gaze estimation dataset~\datasetname, featuring large variation in head pose and gaze direction, high-resolution imagery, varied subject appearance, systematic illumination conditions, as well as accurate ground-truth gaze vectors. Evaluation using a baseline method shows that training on~\datasetname significantly improves robustness towards variation in head pose and gaze direction compared to existing datasets, adding a very valuable resource for future work on gaze estimation.
In addition, we propose a standardized experimental protocol and evaluation framework that will be made available via the benchmark website alongside the dataset, allowing for fair comparison of gaze estimation algorithms on neutral ground.  
\section*{Acknowledgements}
\begin{wrapfigure}{r}{0.3\columnwidth}
    \raggedleft
    \vspace{-4mm}
    \includegraphics[width=0.3\columnwidth]{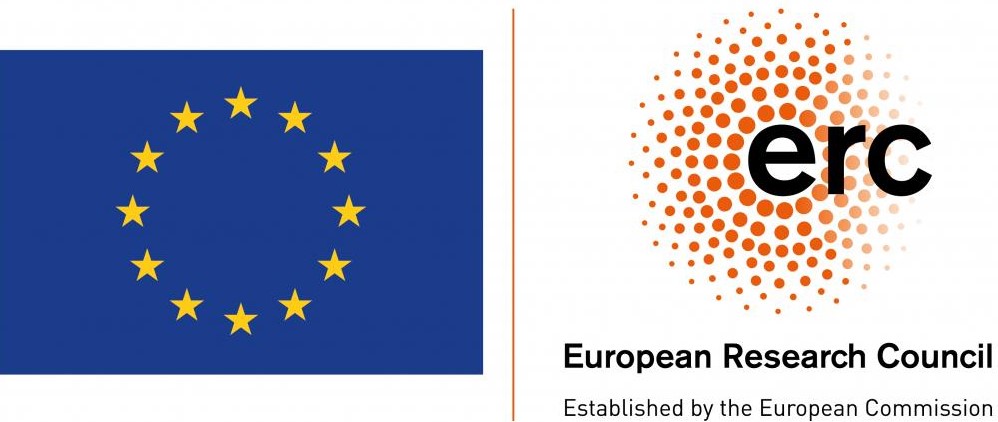}
\end{wrapfigure}
We thank the participants of our dataset for their contributions, our reviewers for helping us improve the paper, and Jan Wezel for helping with the hardware setup.
This project has received funding from the European Research Council (ERC) under the European Union’s Horizon 2020 research and innovation programme grant agreement No. StG-2016-717054.
 \clearpage
\bibliographystyle{splncs04}
\bibliography{reference}
\end{document}